\documentclass[british,sigconf,nonacm]{acmart}
\usepackage[T1]{fontenc}
\usepackage[utf8]{inputenc}
\usepackage{graphicx}

\makeatletter
\usepackage{array}
\AtBeginDocument{\theoremstyle{acmdefinition}\newtheorem{remark}[theorem]{Remark}}
\newfloat{listing}{htbp}{lol}
\floatname{listing}{Listing}
\makeatother

\usepackage{babel}
\begin{document}
\title{Inheritance as an Admission Test for LLM Prompt Merge Operators}
\author{Zan-Kai Chong}
\affiliation{\department{School of Science and Technology}\institution{Kwansei Gakuin University}\city{Sanda}\country{Japan}}
\email{zankai@titanlab.org}
\author{Hiroyuki Ohsaki}
\affiliation{\department{School of Science and Technology}\institution{Kwansei Gakuin University}\city{Sanda}\country{Japan}}
\email{ohsaki@kwansei.ac.jp}
\author{Bryan Ng}
\affiliation{\department{School of Engineering \& Computer Science}\institution{Victoria University of Wellington}\city{Wellington}\country{New Zealand}}
\email{ckbryan@hotmail.com}
\begin{abstract}
Large language model (LLM) agents are often improved by iterating
until a fixed verifier is done. Once that verifier saturates, the
loop has nothing left to climb. We instead score agents against the
current population, and a frozen panel records whether anyone got
better. The panel mean can rise inside the seeds by elimination alone.
We therefore count a run as training only when its final mean beats
the best generation-zero member. A merge is admitted only when the
child moves toward the donor, past the child the same operator produces
at zero dose, and the lower confidence bound on that movement reaches
half the requested dose. That is the admission test. A detectable
shortfall rejects the operator, and the test is otherwise undetermined.
We keep realised fitness $\phi$ as the selector and reference performance
$\theta$ as the unused ruler. Every measurement uses gemini-3.1-flash-lite
on one negotiation task, and no fixed-oracle arm was run. In the open
loop, tactic append grafts short donor rules and is admitted. Free
rewrite moves only a third of the requested dose and is rejected.
Template append pastes raw donor replies into the child. Its movement
clears the bound on the first probe but not on the re-probe, and it
is left undetermined. A held-out re-probe on the five situations the
merge never saw confirms that both appends move, and it does not rescue
free rewrite. Neither append rises with the dose. Constrained rewrite
is admitted in the open loop and was not run in the closed loop, which
changes only the operator. The population, the scores and the task
stay fixed. No free-rewrite run beats the best generation-zero member.
Every tactic-append run and every template-append run does. Sampled
loop children keep that split. Free rewrite moves no further toward
the donor, and both appends still clear the bound, by less than in
the open loop.
\end{abstract}
\maketitle

\section{Introduction}\label{sec:Introduction}

Large language model (LLM) agents are usually improved by iterating
until a fixed verifier is satisfied \cite{shinn2023reflexionlanguageagentsverbal},
and the same shape governs fine-tuning and human feedback \cite{ouyang2022training}.
Once the verifier is satisfied, the loop has nothing left to climb.
We therefore score each agent against opponents drawn from the current
population, and no weights are updated. Realised fitness $\phi$ is
then relative. It ranks the population at every generation, but it
cannot certify progress, because a population in which every agent
improves still faces only itself. We read progress instead on reference
performance $\theta$ against a panel frozen at initialisation, and
$\phi$ stays the selector. That decoupling is the setting of this
paper, not its contribution.

The setting has a weakness, and that weakness motivates the paper.
The panel mean can rise inside the seeds by elimination alone, because
selection keeps the better seeds and discards the worse. A climb therefore
does not show that children inherit anything from their donors. Whether
they do is a property of the merge operator, and that operator is
what we test.

The objects are concrete. Each agent is one frozen LLM with its own
system prompt, and the population is a set of such prompts. A merge
takes a recipient prompt, a behavioural trace of a donor and a requested
dose $\beta$, the fraction of the donor the child is asked to inherit,
and it returns a child prompt. We study four such operators (Table
\ref{tab:definitions}). Two rewrite the recipient and two append
to it. Their effect is read on a behaviour profile, the frequencies
with which a prompt claims each issue of a negotiation game, or accepts,
on a fixed battery of scripted situations.

We ask two questions.
\begin{description}
\item [{RQ1.}] Given a requested dose $\beta$, does an operator move the
child toward the donor, past the child the same operator produces
at $\beta=0$, by the requested dose? We call that the admission test.
An operator is admitted when the lower confidence bound on that movement
reaches half of $\beta$, it is rejected when a shortfall below $\beta$
is detected, and it is otherwise undetermined.
\item [{RQ2.}] Does the open-loop verdict predict the closed loop? When
only the operator changes and the population, the scores and the task
stay fixed, do admitted operators carry the population past the best
generation-zero member, and do rejected ones stay inside the seeds?
\end{description}
The scope is deliberately narrow. We measure transmission, not efficiency.
No fixed-oracle arm was run, and we make no claim that competitive
scoring trains faster than a fixed verifier. Every measurement uses
gemini-3.1-flash-lite on one negotiation task.

We offer two contributions. The first is the admission test of RQ1.
It scores a merge against the operator's own zero-dose child rather
than against the recipient, and that baseline is what separates inherited
movement from paraphrase drift. It admits on a confidence bound rather
than on a failure to detect a shortfall. The second is the closed-loop
protocol of RQ2. A run counts as training only when its final mean
beats the best generation-zero member, which elimination alone cannot
do, and sampled loop children are re-tested with the same admission
test.

\section{Related Work}\label{sec:Related-Work}

The trajectory of LLM systems has run from the static model to the
tool-using agent \cite{yao2022react} and cooperating multi-agent
systems \cite{wu2023autogenenablingnextgenllm}. Unlike those orchestration
frameworks, our agents never divide a task between them, and the output
is a population rather than a workflow. Closest to our work are methods
that optimise a prompt or an agent program against a task metric.
Some use textual gradients \cite{pryzant2023automaticpromptoptimizationgradient}
or an LLM acting as the optimiser or compiler \cite{yang2024largelanguagemodelsoptimizers}.
Others run an evolutionary search over prompt strings \cite{guo2023evoprompt},
including language-model crossover \cite{meyerson2024lmx} and self-referential
prompt evolution \cite{fernando2023promptbreeder}. Those methods
use the parents as in-context examples and sample a new complete prompt.
They do not keep the recipient and append the donor.

By contrast, each merge here takes a recipient prompt, a donor and
a requested dose $\beta$, the fraction of the donor the child is
asked to inherit. We measure that family in Section \ref{sec:LLM-Test}
against its own $\beta=0$ child, the child the same operator produces
at zero dose with the donor still in view, and we contrast it with
a recipient-preserving graft. Headless-chicken crossover replaces
one parent with a random individual \cite{angeline1997headless},
but our control keeps both parents and sets the dose to zero. The
graft itself has precedent. ExpeL and Agent Workflow Memory distil
rules from experience and append them to a prompt \cite{zhao2024expel,wang2024awm}.
Those rules come from the agent's own history, but our donor is a
rival's trace. What is new here is the audit of that graft.

The same neighbourhood now includes later LLM evolution. AlphaEvolve
\cite{novikov2025alphaevolve} edits an existing program, and GEPA
\cite{agrawal2025gepa} edits an existing prompt in light of its rollouts.
ShinkaEvolve and the Darwin Gödel Machine \cite{lange2025shinka,zhang2025darwin}
evolve programs against a task metric, and EvoAgent \cite{yuan2025evoagent}
evolves a specialist into a multi-agent system by mutation and crossover
of agent settings. None of them scores the change against the operator's
own zero-dose child.

The setting inherits from classical evolutionary computation. Competitive
coevolution \cite{hillis1990coevolving,rosin1997coevolution} and
its reinforcement-learning counterpart self-play \cite{silver2018general}
induce the frequency-dependent dynamics we analyse \cite{hofbauer1998evolutionary}.
More recently, foundation-model self-play has brought these dynamics
to LLM-driven evolution \cite{dharna2025fmsp}. That method freezes
the model as well, but it uses the model as a code-level search operator
that writes a new program from the opponent and the match result.
We instead keep the model as the player, and the genome is a prompt
that must inherit. That tradition also supplies our measurement problem
and its pathologies of disengagement, cycling and mediocre stable
states \cite{watson2001coevolutionary}. The frozen panel of Section
\ref{sub:population-fitness-rank} is the simplest of its archives
\cite{lanctot2017psro}.

The merge itself also has classical counterparts. Geometric semantic
crossover places the child on the segment between its parents \cite{moraglio2012geometric}.
We call that placement a blend. A blend inherits exactly the requested
dose, but it never leaves the hull of the seeds. Open-ended and quality-diversity
search \cite{lehman2011novelty} instead seeks the variation a blend
cannot supply, a child outside the hull of its seeds. Remark \ref{rem:ceiling}
distinguishes that hull escape from inherited change.

\section{A Coevolutionary Training Model}\label{sec:Coevolution-Model}

We model competitive survival as a steady-state coevolutionary process
over a fixed population of $N$ agents. Each generation adds $k$
children by a merge operator and removes $k$ agents by rank-proportional
elimination, returning the population to $N$. No weights are updated,
and evolution acts on strategies alone. An agent is scored against
its current rivals rather than a fixed oracle.

\subsection{Population, Competitive Fitness, and Reference Performance}\label{sub:population-fitness-rank}

Let the population at generation $t$ be $P_{t}=\{a_{1},\dots,a_{N}\}$
of fixed size $N$. Each agent carries a strategy, its prompt and
configuration, in a space $\mathcal{A}$. An embedding $\psi:\mathcal{A}\to\mathbb{R}^{d}$
(in Section \ref{sec:LLM-Test}, a behaviour profile) is an instrument
for measuring transmission, and convexity of $\mathcal{A}$ is not
assumed. A competitive task is a two-party interaction with match
function $m(a_{i},a_{j})\in[0,1]$, the expected score agent $i$
obtains against agent $j$. The interaction need not be zero-sum.

An agent's realised fitness is its expected match score against opponents
drawn from the current population,

\begin{equation}
\phi_{i}(P_{t})=\frac{1}{|O_{i}|}\sum_{j\in O_{i}}m(a_{i},a_{j}),\qquad O_{i}\subseteq P_{t}\setminus\{a_{i}\}.\label{eq:fitness}
\end{equation}
Here, $O_{i}$ is a fixed-size set of opponents drawn uniformly without
replacement, independently of the strategies. Because its opponents
are population members, $\phi_{i}$ depends on which strategies its
rivals currently use, and the same strategy scores differently as
that mix shifts.

Because $\phi_{i}$ is relative, it cannot by itself certify progress.
If every agent improves, all still face one another and the mean is
unmoved. We therefore fix a reference panel $B$, at initialisation
and never subject to selection, and define agent $i$'s reference
performance,

\begin{equation}
\theta_{i}=\frac{1}{|B|}\sum_{b\in B}m(a_{i},b).\label{eq:theta}
\end{equation}

Since $B$ never changes, $\theta_{i}$ is an external ruler, the
counterpart of the coevolutionary master tournament \cite{nolfi2025global}.
Progress is the mean reference performance $\bar{\theta}_{t}=\frac{1}{N}\sum_{i}\theta_{i}$.
Finally, sorting agents by $\phi_{i}(P_{t})$ assigns each a rank
$r_{i}\in\{1,\dots,N\}$, with $r_{i}=1$ the fittest.

\subsection{Reproduction Operator}\label{sub:reproduction}

At each generation, $k$ recipients are drawn uniformly with replacement
from the original $N$ scored agents. For each recipient $b$, a donor
$d$ is drawn by the rank-Boltzmann rule (A3) below. The recipient
then requests a dose $\beta\in[0,\beta_{\max}]$, a control in the
prompt rather than a geometric weight, and a child is spawned by a
(possibly stochastic) merge operator $T$, $a_{\mathrm{clone}}\sim T(a_{b},a_{d},\beta)$.
A behavioural trace of the donor may also enter $T$ as a prompt ingredient.
The theory itself uses $T$ only through two derived quantities, the
fidelity and the residual defined next, identified on LLM merge operators
in Section \ref{sec:LLM-Test}.

\begin{figure}[t]
\begin{centering}
\includegraphics[width=1\columnwidth]{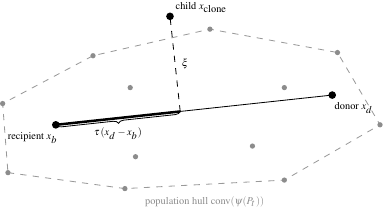}
\par\end{centering}
\caption{The decomposition of Eq. \ref{eq:transmission} in the embedding.
Points are population members $\psi(a)$. The child $x_{\mathrm{clone}}$
is the recipient $x_{b}$ plus a fraction $\tau$ of the donor increment,
the projection of Eq. \ref{eq:fidelity}, plus the orthogonal residual
$\xi$. The dashed polygon is the hull of $\psi(P_{t})$, which the
residual may leave.}\label{fig:instrument}
\end{figure}

Write $x=\psi(a)$ for the embedding of a strategy. The child is the
recipient, plus a fraction $\tau$ of the donor increment (what was
inherited), plus an orthogonal residual $\xi$. Formally, any realised
child with $x_{d}\neq x_{b}$ decomposes uniquely as

\begin{equation}
x_{\mathrm{clone}}=x_{b}+\tau(x_{d}-x_{b})+\xi.\label{eq:transmission}
\end{equation}
The realised fidelity

\begin{equation}
\tau=\frac{(x_{\mathrm{clone}}-x_{b})^{\top}(x_{d}-x_{b})}{\lVert x_{d}-x_{b}\rVert^{2}}\label{eq:fidelity}
\end{equation}
is the coefficient that makes the residual $\xi$ orthogonal to the
donor-recipient difference (Fig. \ref{fig:instrument}). The fidelity
of an operator at a requested $\beta$, $\tau(\beta)$, is its expectation
under $T$. If reference performance is affine in the embedding, the
same split holds on the panel, $\theta_{\mathrm{clone}}=\theta_{b}+\tau(\theta_{d}-\theta_{b})+\varepsilon$,
with $\mu=\mathbb{E}[\varepsilon]$ the operator bias. The expectation
is under $T$ for a given pair, as for $\tau(\beta)$. Orthogonality
fixes each realised $\xi$ but does not make its mean vanish, and
$\mu$ is the panel reading of that mean shift. The realised residual
$\xi$, not its mean, may leave the convex hull of the current population.

We adopt four modelling assumptions. (A1) A child satisfies Eq. \ref{eq:transmission}
in the embedding, with fidelity $\tau(\beta)$ and orthogonal residual
$\xi$. Both are properties of the operator family, identified empirically.
(A2) A merge leaves the recipient unchanged, and the operator bias
$\mu$ lands in the child alone. That bias does not vary with the
dose, and it is read at $\beta=0$, where a rewrite still paraphrases
while an append copies. (A3) The donor is drawn with probability $\Pr(d=i)\propto e^{-\eta\,(r_{i}-1)}$,
decaying in realised-fitness rank with inverse temperature $\eta>0$.
(A4) Conditional on $(a_{b},a_{d},\beta)$, the residual $\xi$ is
independent across children and generations. It need not be independent
of $\beta$, and the $\beta=0$ rewrite is the baseline law of $\xi$.

Requested $\beta$ and fidelity $\tau$ are different objects. The
requested dose follows the rule

\begin{equation}
\beta^{*}(\phi_{b},\phi_{d})=\min\!\left(\beta_{\max},\ \max\!\left(0,\ \frac{\phi_{d}-\phi_{b}}{\kappa}\right)\right).\label{eq:blend}
\end{equation}

That rule requests a dose only from a fitter donor and raises it with
the fitness gap, up to the cap $\beta_{\max}$. The scale $\kappa$
is the gap at which the requested dose would reach one. It reads $\beta$
as the fraction inherited, which holds for a blend, where $\tau(\beta)=\beta$.
If $\tau$ is instead flat in $\beta$, as the appends are, the same
rule is a misspecified policy. If $\tau$ is a small fraction of $\beta$,
as free rewrite is, the rule under-delivers at every dose.

\subsection{Population Dynamics}\label{sub:population-dynamics}

Children enter unscored, hence immune to elimination in the current
generation. Each generation then removes $k$ agents sequentially
without replacement, with weight equal to rank. The least-fit agent
therefore carries the greatest weight, and it is the most likely to
leave. What that removal moves is the mean reference performance $\bar{\theta}_{t}$.

Since the recipient is drawn uniformly, $\mathbb{E}[\theta_{b}]=\bar{\theta}_{t}$.
One generation of $k$ merges and $k$ removals then changes the panel
mean by

\begin{equation}
\Delta\bar{\theta}_{t}=\frac{k}{N}\Bigl[\,\mathbb{E}[\tau(\beta)(\theta_{d}-\theta_{b})]+\mu-\tfrac{2}{N+1}\mathrm{Cov}_{i}(r_{i},\theta_{i})\Bigr].\label{eq:mean-change}
\end{equation}

The first term is the fidelity term, the second is the bias term,
and the third is the selection term, which is one rank-weighted removal
$\mathbb{E}[\theta_{\mathrm{elim}}]-\bar{\theta}_{t}$ written out.
Because rank one is the fittest, the covariance is negative whenever
reference performance tracks realised fitness, and the selection term
then lifts the mean even when no child inherits anything. The decomposition
is exact for $k=1$ and holds to leading order in $k/N$ otherwise
\cite{price1970selection}. For a convex blend, the fidelity term
becomes $\mathbb{E}[\beta^{*}(\phi_{b},\phi_{d})\,(\theta_{d}-\theta_{b})]$
and $\mu=0$.

\subsection{Fidelity and the Operator Envelope}\label{sub:diversity-dynamics}

Section \ref{sub:population-dynamics} accounts for one generation
of change. This subsection asks two questions of the operator. First,
what does selection achieve when the child does not respond to dose?
Second, how far can any operator carry the population from its seeds?
\begin{proposition}[Response to selection is gated by dose]
\label{prop:fidelity-gating}Let $z$ be any trait that is affine
in the embedding, $\theta$ among them when reference performance
is. The one-generation response of the mean to selection on the donor
increment $z_{d}-z_{b}$ is the fidelity term. If $\tau(\beta)$ does
not vary with the requested dose, the increment is inherited at a
rate the dose cannot set, and raising the dose with the donor's advantage
does no extra work. The extreme case $\tau(\beta)=0$ is included.
Then the increment is not inherited at all, however large the donor's
advantage.
\end{proposition}
\begin{remark}[Ceiling is the operator envelope]
\label{rem:ceiling}Every later strategy lies in the closed support
of the finite iterates of $T$ starting from $P_{0}$. If $\theta$
is continuous, $\sup_{i}\theta_{i}(P_{t})$ cannot exceed the supremum
of $\theta$ on that support. In general this supremum need not equal
$\max_{i}\theta_{i}(P_{0})$. Under a blend with $\tau(\beta)=\beta$
and vanishing residual, the embedding stays in the hull of $P_{0}$.
When reference performance is affine in the embedding, no agent surpasses
the best member of $P_{0}$. That is why a noiseless blend cannot
beat the best seed. Leaving the parent line is necessary to clear
the best seed, and it is not sufficient. A residual that is mere noise
can also clear the best seed, hence a large unsigned residual does
not by itself admit a merge.
\end{remark}

\section{Identifying Fidelity on LLM Merge Operators}\label{sec:LLM-Test}

This section answers RQ1. It measures a single merge in simulation,
with no ranking. It asks whether the child moved toward the donor,
and whether it left the parent line.

\subsection{Task and Operators}\label{sub:llm-setup}

In this simulation, the agents are gemini-3.1-flash-lite models that
differ only in their system prompts. Twelve seed prompts of one or
two sentences form the generation-zero population, and they span hard-liner,
trading, mirroring, splitting and conceding styles.

Those agents play a structured variant of the Deal or No Deal task
\cite{lewis2017deal}. The game has three indivisible issues, a patent,
a territory and a staffing decision. Each issue goes wholly to one
side. To reach a deal, the two agents exchange six messages rather
than free dialogue. Each message is a proposal of an allocation or
an acceptance of the offer on the table. Each side holds private values
for the three issues. Both sides score zero if they never agree, and
an agent's score is otherwise the share of its total value that the
deal captures.

A child is made from a recipient prompt and a donor trace. The merge
model is the same gemini-3.1-flash-lite, but it never sees the donor's
prompt. It instead sees three of the donor's replies to the eight
scripted situations of Section \ref{sub:llm-tests}. For that merge,
a requested dose $\beta$ is a percentage written into the call, or
a count of grafted lines for the appends, and it is never a geometric
weight.

Table \ref{tab:definitions} defines the four operators. Free rewrite,
tactic append and template append later enter the loop of Section
\ref{sec:Loop}. Among them, tactic append keeps the recipient and
grafts short donor rules from a merge-model call.

Free rewrite, by contrast, regenerates the whole prompt, and none
of the recipient's wording is required to survive. Template append
keeps the recipient verbatim and grafts raw donor-trace lines under
a fixed wrapper, and it makes no model call. Constrained rewrite keeps
unreplaced recipient text, but it is measured only in the open loop.
Hence at $\beta=0$ append does nothing and the child is the recipient,
whereas free rewrite at $\beta=0$ still paraphrases.

\begin{table*}[t]
\caption{The four merge operators. Each takes a recipient prompt, a donor trace and a requested dose $\beta$. The rewrites pass $\beta$ to the model as a requested percentage. Template append uses $\beta$ as a count of grafted lines and makes no model call. Tactic append calls the model for short rules, then appends that count mechanically.}
\label{tab:definitions}
\centering
\small
\begin{tabular}{llll}
\hline
operator & model call & donor material & $\beta=0$ child\\
\hline
free rewrite & regenerates the whole prompt & trace, as context & paraphrase\\
constrained rewrite & rewrites, keeping unreplaced text verbatim & trace, as context & original, if honoured\\
tactic append & abstracts the trace into short rules only & rules, appended & recipient\\
template append & none & raw trace lines, appended & recipient\\
\hline
\end{tabular}
\end{table*}

\subsection{Two Tests on Single Merges}\label{sub:llm-tests}

Behaviour is measured on a fixed battery, not against a live opponent.
Each strategy answers eight scripted situations. Each situation is
one turn of the game, at the opening, the midgame or the final message,
and the opponent's earlier messages are written into the script rather
than generated by a rival (Listing \ref{lst:situation}). Because
the values, history and turn index are identical for every strategy,
any difference between two profiles is due to the strategies and to
sampling alone.

\begin{listing}
\small
\begin{tabular}{|p{0.95\columnwidth}|}
\hline
\textbf{System.} \emph{[the strategy's own prompt]} You are negotiating over three indivisible issues: patent, territory, staffing. Each must go entirely to A or to B. You are side A. Your private values are patent 50, territory 30, staffing 20. This is message 5 of 6; 1 message remains after yours. Reply with \texttt{propose} or \texttt{accept} and an allocation. If no agreement is reached by the last message both sides score zero.\\[0.3em]
\textbf{User.} Negotiation so far:\\
A: propose patent A, territory A, staffing B. \emph{my terms}\\
B: propose patent B, territory B, staffing A. \emph{staffing is all I can give}\\
A: propose patent A, territory B, staffing A. \emph{I need the patent}\\
B: propose patent B, territory B, staffing A. \emph{unchanged, take it or leave it}\\
Your move.\\
\hline
\end{tabular}
\caption{One of the eight scripted situations, at side A's final message. The strategy's prompt is prepended to the system message, and the four B lines are fixed text. Accepting here scores 20, and any other move scores zero.}
\label{lst:situation}
\end{listing}

On that battery, each situation is sampled 24 times at temperature
0.8. Each reply is a proposed allocation or an acceptance, and four
binary features record whether the reply claims each issue for side
A and whether it accepts. Their frequencies are the behaviour profile,
the embedding $\psi$ of Section \ref{sec:Coevolution-Model} under
Euclidean distance. The whole battery is sampled twice for every strategy,
and its profile is the mean of those two passes. The gap between the
passes measures the sampling noise, and the residual noise of the
mean is 0.06 of the median recipient-donor separation, which sets
the detection limit for a dose.

The donor trace is drawn from the first three situations, and three
of the eight scoring prompts carry the donor's own reply. Every child
is therefore re-probed on the same battery, in two further passes,
and its movement is read on two blocks of the profile, the full battery
and the five held-out situations the merge never saw. The noise floor
and the convex null are recomputed on each block.

To read that movement, we draw a line from the recipient's profile
to the donor's, as in Fig. \ref{fig:instrument}. Place the recipient
at 0 and the donor at 1 on that line. The child's position on that
line is a single number $s$. Here $s=0$ means the child still acts
like the recipient, and $s=1$ means it now acts like the donor. That
number is the realised $\tau$ of Eq. \ref{eq:fidelity}. If $T$
were a convex blend, the child would lie on that recipient-donor line.
The noise floor in the child and in the two endpoints would still
displace it off that line by a known amount. We call that amount the
convex null and report off-segment residuals relative to it.

\begin{figure}[t]
\begin{centering}
\includegraphics[width=1\columnwidth]{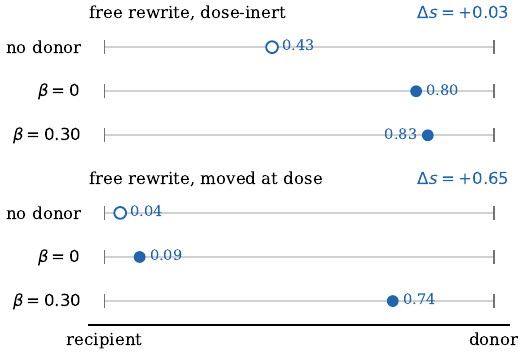}
\par\end{centering}
\caption{Two free-rewrite pairs, with the recipient at 0 and the donor at 1.
The filled $\beta=0$ marker is the zero-dose child that serves as
the baseline, and the hollow marker is the no-donor control of Table
\ref{tab:operators}, the same rewrite instruction with the rival
block removed.}\label{fig:s-lines}
\end{figure}

Four operators were run on sixteen recipient-donor pairs with eleven
distinct donors, at doses $\beta\in\{0,0.15,0.30\}$, with one child
per operator, pair and dose, as illustrated in Fig. \ref{fig:s-lines}.
The increment $\Delta s=s(\beta)-s(0)$ is the extra movement that
the requested dose caused beyond that operator's own zero-dose child.
Table \ref{tab:operators} reports $s(0)$ as a median over the sixteen
pairs rather than a mean, and that median is our estimate of the operator
bias $\mu$ of Section \ref{sub:reproduction}, read in the embedding
and chosen for robustness over so few pairs. Single pairs overshoot
the donor or move backwards, and a mean of sixteen values follows
them, whereas the rank-based tests below read the median. Every $\Delta s$
reported below is therefore a median as well. In Fig. \ref{fig:s-lines}
the dose moves the child in one row and not in the other, and both
rows take the zero-dose child as their baseline.

That baseline matters because a free rewrite requested at $\beta=0$
still paraphrases the recipient, and a paraphrase can drift toward
the donor with no donor in view (the no-donor row of Table \ref{tab:operators}).
Whether the trace is in view makes no systematic difference at zero
dose. The with-trace child sits beyond the no-donor child on eight
of sixteen pairs, and the median gap is well inside the noise floor,
though single pairs such as the dose-inert row of Fig. \ref{fig:s-lines}
differ widely. The drift is the paraphrase itself. We therefore read
each child against the same operator's own $\beta=0$ child.

The admission test asks whether $\Delta s$ reaches the requested
dose. Movement and shortfall are each read by a one-sided Wilcoxon
signed-rank test over the sixteen pairs, and the operator is rejected
when that test detects a shortfall. But a shortfall that goes undetected
is not a dose delivered, because sixteen pairs with a spread of 0.3
cannot tell $\Delta s=0.30$ from $\Delta s=0.15$. Admission therefore
rests on a bound rather than on a non-rejection. We take the one-sided
95\% lower confidence bound on $\Delta s$ that the signed-rank test
implies, the Hodges-Lehmann bound, and we admit the operator only
when that bound reaches half the requested dose. Half the reference
effect is the common non-inferiority margin, and it sits well above
the detection limit of 0.06. An operator that is neither rejected
nor admitted is undetermined, which is a statement about the sample
rather than about the operator. The full battery decides, and an admission
stands only if the re-probe of the same children clears the bound
as well. The held-out block then asks whether the movement survives
once the three seen situations are removed, and it cannot rescue an
operator the full battery rejected.

\subsection{Admitting the Operators}\label{sub:llm-operators}

Section \ref{sub:llm-tests} defined the extra movement past each
operator's own zero-dose child. This subsection applies that comparison
to all sixteen pairs, and it also records whether the child leaves
the parent line.

Table \ref{tab:operators} gives the admission test for each operator.
Tactic append moves past its own zero-dose child by a median $\Delta s$
of $+0.33$ at both doses. At $\beta=0.30$ the lower bound on that
movement is $+0.21$ on the first probe and $+0.19$ on the re-probe,
both above the half-dose bar of 0.15, and on the held-out block fifteen
of sixteen pairs move forward ($p<0.001$) with a bound of $+0.15$.
It is admitted. Constrained rewrite moves by $+0.40$ with bounds
of $+0.23$ and $+0.21$, and it is admitted as well, but it was not
run in the closed loop. Free rewrite, by contrast, moves by $+0.10$
at $\beta=0.30$, a third of the requested dose. The Wilcoxon test
finds that movement greater than zero, but on the full battery it
falls significantly short of the dose ($p=0.03$ on the first probe
and $p=0.04$ on the re-probe), and its bound of $+0.05$ is a third
of the bar. It is rejected. On the held-out block alone the shortfall
is not detectable, but that block cannot rescue a full-battery rejection.

Template append, in turn, moves by a median $\Delta s$ of $+0.35$
at $\beta=0.30$ on the first probe, with a bound of $+0.17$ that
clears the bar. But the movement is suspect, because the pasted lines
are the donor's own replies on three of the eight scoring situations,
and the child could be moving by quoting the answer rather than by
changing its behaviour. The re-probe settles the quoting question.
On the five situations the merge never saw, template append still
moves by $+0.18$ with fourteen of sixteen pairs positive ($p<0.001$),
as far as on the three seen situations ($+0.16$). The movement is
real. But the re-probe's full-battery bound is $+0.11$, below the
bar, and the admission does not stand. Template append is therefore
undetermined. It is not rejected, since no shortfall is detected ($p=0.68$),
and sixteen pairs cannot certify its dose.

Two caveats stay. Re-probing the same children returned a full-battery
median of $+0.23$ rather than $+0.35$, while every other operator's
median returned within 0.04. Six pairs barely move and the rest move
a lot. That split is what leaves the verdict undetermined, because
the bound follows the weak pairs. The second caveat is shared with
tactic append, because neither append tracks the requested dose. Tactic
append moves by the same $+0.33$ at both doses, and template append
moves further at $\beta=0.15$ than at $\beta=0.30$ on every block.
For the appends the dose is a count of grafted lines, and a second
line adds little once the first has landed.

We record whether the child leaves the parent line. That departure
is the off-segment residual relative to the convex null, and it is
the realised $\xi$ of Eq. \ref{eq:transmission}. At the requested
dose every operator lands 5.6 to 7.9 times the convex null off the
segment, while a noiseless blend stays on the line. That residual
does not select among these four.

\begin{table*}[t]
\caption{Transmission and drift for the operators of Table~\ref{tab:definitions} over sixteen recipient-donor pairs. $s(0)$: median position of the operator's own $\beta=0$ child. $\Delta s$: median movement beyond that child at each dose, on the first probe of the full battery. Bound: one-sided 95\% lower confidence bound on $\Delta s$ at $\beta=0.30$, on the first probe and on the re-probe of the same children. Ratio: off-segment residual over the convex null at $\beta=0.30$. Held-out: $\Delta s$ at $\beta=0.30$ on the five held-out situations, and its bound. Verdict: admitted when both full-battery bounds reach half the dose, $0.15$, rejected when a shortfall below the dose is detected, and otherwise undetermined.}
\label{tab:operators}
\centering
\small
\begin{tabular}{lccccccccc}
\hline
operator & $s(0)$ & \begin{tabular}[b]{@{}c@{}}$\Delta s$\\($0.15$)\end{tabular} & \begin{tabular}[b]{@{}c@{}}$\Delta s$\\($0.30$)\end{tabular} & \begin{tabular}[b]{@{}c@{}}bound\\(probe 1)\end{tabular} & \begin{tabular}[b]{@{}c@{}}bound\\(re-probe)\end{tabular} & ratio & \begin{tabular}[b]{@{}c@{}}held-out\\$\Delta s$\end{tabular} & \begin{tabular}[b]{@{}c@{}}held-out\\bound\end{tabular} & verdict\\
\hline
tactic append & $0$ & $+0.33$ & $+0.33$ & $+0.21$ & $+0.19$ & 7.9 & $+0.23$ & $+0.15$ & admitted\\
constrained rewrite & $0$ & $+0.15$ & $+0.40$ & $+0.23$ & $+0.21$ & 5.6 & $+0.29$ & $+0.16$ & admitted\\
template append & $0$ & $+0.41$ & $+0.35$ & $+0.17$ & $+0.11$ & 7.6 & $+0.18$ & $+0.10$ & undetermined\\
free rewrite & $0.30$ & $+0.06$ & $+0.10$ & $+0.05$ & $+0.06$ & 6.5 & $+0.14$ & $+0.03$ & rejected\\
free rewrite, no donor & $0.39$ & -- & -- & -- & -- & 7.9 & -- & -- & --\\
\hline
\end{tabular}
\end{table*}

Because free rewrite's failure belongs to the operator, the loop needs
transmission (Proposition \ref{prop:fidelity-gating}), not hull membership
(Remark \ref{rem:ceiling}). The failure also reaches beyond this
paper. PromptBreeder, EvoPrompt and language-model crossover \cite{fernando2023promptbreeder,guo2023evoprompt,meyerson2024lmx}
recombine by operators of the same kind, which sample a new complete
prompt from the parents' text. If those operators transmit as ours
does, their selection step chooses parents whose behaviour the child
does not reliably inherit, and the operator supplies fresh variation
in their place. An audit of LLM evolution in equation discovery points
the same way, finding instructed crossover no better than fresh sampling
\cite{li2026dictionaries}. Section \ref{sec:Loop} re-measures $\Delta s$
on sampled loop children.

\section{The Closed Loop on LLM Agents}\label{sec:Loop}

This section answers RQ2, whether the open-loop verdict predicts the
closed loop. Section \ref{sec:LLM-Test} admitted tactic append and
rejected free rewrite. Template append makes no model call, and it
was left undetermined, with a movement the held-out block confirms
and a bound the re-probe does not clear. This section closes the loop,
and only the merge operator $T$ changes. The task is the negotiation
game again, with the same model and the same twelve seed prompts.

\begin{table*}[t]
\caption{Open-loop verdict (Table~\ref{tab:operators}, $\beta=0.30$) beside closed-loop outcome. $\Delta s$ and ratio: as in Table~\ref{tab:operators}. Verdict: the open-loop verdict. Loop $\Delta s$: median $\Delta s$ of sampled closed-loop children on the same battery. Runs past best seed: runs whose final mean beats the pooled bar of 0.771. Child increment: mean panel gain from recipient to child.}
\label{tab:bridge}
\centering
\small
\begin{tabular}{lccccrr}
\hline
 & \multicolumn{3}{c}{Section~\ref{sec:LLM-Test} (open loop)} & \multicolumn{3}{c}{Section~\ref{sec:Loop} (closed loop)}\\
\cline{2-4}\cline{5-7}
operator & $\Delta s$ & ratio & verdict & loop $\Delta s$ & runs past best seed & child increment\\
\hline
tactic append & $+0.33$ & 7.9 & admitted & $+0.19$ & 3/3 & $+0.028$\\
template append & $+0.35$ & 7.6 & undetermined & $+0.17$ & 3/3 & $+0.011$\\
free rewrite & $+0.10$ & 6.5 & rejected & $+0.00$ & 0/3 & $-0.001$\\
\hline
\end{tabular}
\end{table*}

The loop runs $N=12$ strategies with turnover $k=2$ for 40 generations,
in three coevolutionary runs. Each agent's realised fitness is its
mean score against five rivals drawn from the other eleven, with matches
sampled at temperature $0.3$. The reference panel is six members
of the frozen generation-zero population, and the population is scored
against it at every other generation, with panel matches sampled at
temperature 0.15. The donor trace is taken on the battery of Section
\ref{sub:llm-tests}. The donor is drawn with $\eta=0.5$ under (A3),
and the requested dose follows Eq. \ref{eq:blend} with $\kappa=0.1$
and $\beta_{\max}=0.3$.

\begin{figure}[t]
\begin{centering}
\includegraphics[width=1\columnwidth]{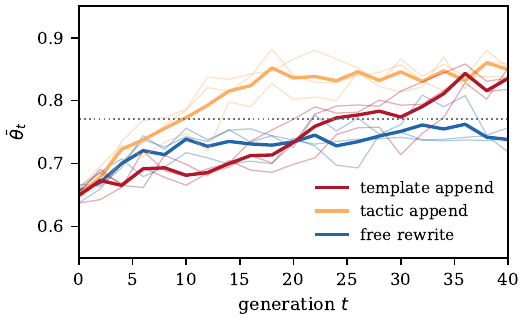}
\par\end{centering}
\caption{Thin curves are each run and the solid curve is their mean, both mean
reference performance $\bar{\theta}_{t}$ on the frozen generation-zero
panel. The dotted line is the pooled best generation-zero score of
0.771, the bar for every run.}\label{fig:loop}
\end{figure}

Fig. \ref{fig:loop} shows that mean reference performance climbs
under all three operators, but we count a run as training only if
its final mean beats the best generation-zero member. Each seed's
generation-zero score in one run rests on six matches, and the largest
of twelve such estimates is noisy. Across the nine runs it ranges
from 0.737 to 0.880 for the same twelve prompts. We therefore pool
each seed's generation-zero scores across the nine runs, 54 matches
per seed, and the best seed then scores 0.771. That is the bar for
every run. Tactic append rises from 0.65 to 0.850, and every run's
final mean clears the bar (0.838, 0.852 and 0.858). Free rewrite rises
from 0.66 to 0.738, and no run clears it (0.757, 0.718 and 0.739).
That climb stays inside the seeds. Template append rises from 0.65
to 0.835, and every run clears the bar (0.836, 0.818 and 0.852). It
pastes raw donor replies, and the held-out block of Section \ref{sec:LLM-Test}
confirms that the movement is real. Against each run's own six-match
bar the counts are two of three, none of three and three of three,
and the one miss is the tactic-append run that drew the 0.880 bar
while its mean rose from 0.652 to 0.838.

The child increment of Table \ref{tab:bridge} is the child's panel
score minus the recipient's, averaged over merges that request a positive
dose. That mean is $+0.028$ under tactic append, $-0.001$ under
free rewrite and $+0.011$ under template append. Tactic append does
not paste raw donor replies, and its children still improve on the
panel. A noiseless blend stays within the hull of the seeds (Remark
\ref{rem:ceiling}), and the append final means do not. The panel
score is a single number, and it can rise toward a strong donor's
without any behaviour being inherited. The admission test is therefore
re-run on sampled loop children with the same eight-situation battery
as Table \ref{tab:operators}. Twelve dosed merges are drawn from
each run, including late generations. A zero-dose append returns its
recipient verbatim, and copies accumulate. By generation 40 between
five and ten of the twelve agents in each append run are copies of
another, against one to three under free rewrite. A pair whose recipient
and donor are the same text has no line to read and is dropped, which
leaves 36 pairs for free rewrite, 24 for tactic append and 31 for
template append.

On that battery tactic append moves past its own zero-dose child by
a median $\Delta s$ of $+0.19$ with twenty of twenty-four pairs
positive ($p<0.001$). The requested dose varies by merge under Eq.
\ref{eq:blend}, hence the bound is taken on $\Delta s-\beta/2$ across
the sampled merges, and for tactic append it is $+0.01$. Template
append moves by $+0.17$ with twenty-five of thirty-one positive ($p<0.001$),
and its bound is $+0.02$, with the same caveats on stability and
dose. Free rewrite moves by $+0.00$ with eighteen of thirty-six positive,
and its shortfall is detectable ($p<0.001$). Both appends therefore
clear the admission bound inside the loop, though by a small margin,
and free rewrite is rejected. The free-rewrite climb is elimination
inside the seeds. Free rewrite's own zero-dose sibling already sits
at $s(0)=+0.23$. The child's raw position on the parent line is then
paraphrase, not dose. The magnitude is smaller than in Table \ref{tab:operators},
where diverse seed pairs at $\beta=0.30$ gave $+0.33$ for tactic
append, $+0.10$ for free rewrite and $+0.35$ for template append.
Table \ref{tab:bridge} sets that verdict beside the closed-loop outcome.

\section{Conclusion}\label{sec:Conclusion}

Iterating against a fixed verifier stops once the verifier is satisfied.
Scoring against the current population keeps a target that moves with
the population, and the frozen panel with the best-seed bar reports
whether that movement is progress. The gain is not efficiency, which
we did not measure against a fixed-oracle arm, but a target that does
not saturate and a reading of progress that elimination cannot fake.

Competitive survival guarantees a ranking at every generation (Section
\ref{sec:Coevolution-Model}), and a frozen panel records whether
the population climbed (Eq. \ref{eq:theta}). A ranking by itself
does not certify progress, and elimination alone can lift the panel
mean without anyone leaving the seeds. Training is therefore read
against the best seed, not against the starting mean. Whether a run
can clear that mark is a property of the operator, not of the tournament.

On RQ1, free rewrite leaves the parent line, but it moves only a third
of the requested dose and is rejected. Tactic append clears the half-dose
bound on both probes and leaves the line without pasting raw donor
replies, and the held-out situations confirm it. Template append makes
no model call and pastes raw donor replies. Its movement is real on
the five situations the merge never saw, but its bound clears the
bar on one probe of the full battery of Section \ref{sec:LLM-Test}
and not on the other, hence it is undetermined, and neither append
rises with the dose. On RQ2, the open-loop verdict predicted the closed
loop for both decided operators that entered it, and the undetermined
operator trained as well. In the closed loop, no free-rewrite run
clears that mark, and every tactic-append run and every template-append
run does. Sampled loop children keep that split, and both appends
clear the admission bound inside the loop while free rewrite is rejected.
Every measurement uses one model on one negotiation task, with sixteen
pairs in the open loop and three runs per operator in the closed loop,
and both appends drive the population toward copies of a few agents.

\bibliographystyle{ACM-Reference-Format}
\bibliography{bibliography}


\begin{thebibliography}{30}

%%% the Web address.

\ifx \showCODEN    \undefined \def \showCODEN     #1{\unskip}     \fi
\ifx \showDOI      \undefined \def \showDOI       #1{#1}\fi
\ifx \showISBNx    \undefined \def \showISBNx     #1{\unskip}     \fi
\ifx \showISBNxiii \undefined \def \showISBNxiii  #1{\unskip}     \fi
\ifx \showISSN     \undefined \def \showISSN      #1{\unskip}     \fi
\ifx \showLCCN     \undefined \def \showLCCN      #1{\unskip}     \fi
\ifx \shownote     \undefined \def \shownote      #1{#1}          \fi
\ifx \showarticletitle \undefined \def \showarticletitle #1{#1}   \fi
\ifx \showURL      \undefined \def \showURL       {\relax}        \fi
% invisible to TeX
\providecommand\bibfield[2]{#2}
\providecommand\bibinfo[2]{#2}
\providecommand\natexlab[1]{#1}
\providecommand\showeprint[2][]{arXiv:#2}

\bibitem[Agrawal et~al\mbox{.}(2025)]%
        {agrawal2025gepa}
\bibfield{author}{\bibinfo{person}{Lakshya~A Agrawal},
  \bibinfo{person}{Shangyin Tan}, \bibinfo{person}{Dilara Soylu},
  \bibinfo{person}{Noah Ziems}, \bibinfo{person}{Rishi Khare},
  \bibinfo{person}{Krista Opsahl-Ong}, \bibinfo{person}{Arnav Singhvi},
  \bibinfo{person}{Herumb Shandilya}, \bibinfo{person}{Michael~J Ryan},
  \bibinfo{person}{Meng Jiang}, \bibinfo{person}{Christopher Potts},
  \bibinfo{person}{Koushik Sen}, \bibinfo{person}{Alexandros~G. Dimakis},
  \bibinfo{person}{Ion Stoica}, \bibinfo{person}{Dan Klein},
  \bibinfo{person}{Matei Zaharia}, {and} \bibinfo{person}{Omar Khattab}.}
  \bibinfo{year}{2025}\natexlab{}.
\newblock \bibinfo{title}{{GEPA}: Reflective Prompt Evolution Can Outperform
  Reinforcement Learning}.
\newblock
\newblock
\showeprint[arxiv]{2507.19457}~[cs.CL]
\newblock
\shownote{arXiv:2507.19457}.


\bibitem[Angeline(1997)]%
        {angeline1997headless}
\bibfield{author}{\bibinfo{person}{Peter~J. Angeline}.}
  \bibinfo{year}{1997}\natexlab{}.
\newblock \showarticletitle{Subtree Crossover: Building Block Engine or
  Macromutation?}. In \bibinfo{booktitle}{\emph{Genetic Programming 1997:
  Proceedings of the Second Annual Conference}}. \bibinfo{publisher}{Morgan
  Kaufmann}, \bibinfo{address}{San Francisco, CA}, \bibinfo{pages}{9--17}.
\newblock


\bibitem[Dharna et~al\mbox{.}(2025)]%
        {dharna2025fmsp}
\bibfield{author}{\bibinfo{person}{Aaron Dharna}, \bibinfo{person}{Cong Lu},
  {and} \bibinfo{person}{Jeff Clune}.} \bibinfo{year}{2025}\natexlab{}.
\newblock \bibinfo{title}{Foundation Model Self-Play: Open-Ended Strategy
  Innovation via Foundation Models}.
\newblock
\newblock
\showeprint[arxiv]{2507.06466}~[cs.LG]
\newblock
\shownote{arXiv:2507.06466}.


\bibitem[Fernando et~al\mbox{.}(2023)]%
        {fernando2023promptbreeder}
\bibfield{author}{\bibinfo{person}{Chrisantha Fernando}, \bibinfo{person}{Dylan
  Banarse}, \bibinfo{person}{Henryk Michalewski}, \bibinfo{person}{Simon
  Osindero}, {and} \bibinfo{person}{Tim Rockt{\"a}schel}.}
  \bibinfo{year}{2023}\natexlab{}.
\newblock \bibinfo{title}{{Promptbreeder}: Self-Referential Self-Improvement
  Via Prompt Evolution}.
\newblock
\newblock
\showeprint[arxiv]{2309.16797}~[cs.CL]
\newblock
\shownote{arXiv:2309.16797}.


\bibitem[Guo et~al\mbox{.}(2023)]%
        {guo2023evoprompt}
\bibfield{author}{\bibinfo{person}{Qingyan Guo}, \bibinfo{person}{Rui Wang},
  \bibinfo{person}{Junliang Guo}, \bibinfo{person}{Yinong Xu}, {and}
  \bibinfo{person}{Bowen Zhou}.} \bibinfo{year}{2023}\natexlab{}.
\newblock \bibinfo{title}{{EvoPrompt}: Connecting Evolutionary Algorithms with
  Prompt Optimization}.
\newblock
\newblock
\showeprint[arxiv]{2309.08532}~[cs.CL]
\newblock
\shownote{arXiv:2309.08532}.


\bibitem[Hillis(1990)]%
        {hillis1990coevolving}
\bibfield{author}{\bibinfo{person}{W.~Daniel Hillis}.}
  \bibinfo{year}{1990}\natexlab{}.
\newblock \showarticletitle{Co-evolving parasites improve simulated evolution
  as an optimization procedure}.
\newblock \bibinfo{journal}{\emph{Physica D: Nonlinear Phenomena}}
  \bibinfo{volume}{42}, \bibinfo{number}{1--3} (\bibinfo{year}{1990}),
  \bibinfo{pages}{228--234}.
\newblock


\bibitem[Hofbauer and Sigmund(1998)]%
        {hofbauer1998evolutionary}
\bibfield{author}{\bibinfo{person}{Josef Hofbauer} {and} \bibinfo{person}{Karl
  Sigmund}.} \bibinfo{year}{1998}\natexlab{}.
\newblock \bibinfo{booktitle}{\emph{Evolutionary Games and Population
  Dynamics}}.
\newblock \bibinfo{publisher}{Cambridge University Press},
  \bibinfo{address}{Cambridge}.
\newblock


\bibitem[Lanctot et~al\mbox{.}(2017)]%
        {lanctot2017psro}
\bibfield{author}{\bibinfo{person}{Marc Lanctot}, \bibinfo{person}{Vinicius
  Zambaldi}, \bibinfo{person}{Audrunas Gruslys}, \bibinfo{person}{Angeliki
  Lazaridou}, \bibinfo{person}{Karl Tuyls}, \bibinfo{person}{Julien
  P{\'e}rolat}, \bibinfo{person}{David Silver}, {and} \bibinfo{person}{Thore
  Graepel}.} \bibinfo{year}{2017}\natexlab{}.
\newblock \showarticletitle{A unified game-theoretic approach to multiagent
  reinforcement learning}. In \bibinfo{booktitle}{\emph{Advances in Neural
  Information Processing Systems}}, Vol.~\bibinfo{volume}{30}.
\newblock


\bibitem[Lange et~al\mbox{.}(2025)]%
        {lange2025shinka}
\bibfield{author}{\bibinfo{person}{Robert~Tjarko Lange}, \bibinfo{person}{Yuki
  Imajuku}, {and} \bibinfo{person}{Edoardo Cetin}.}
  \bibinfo{year}{2025}\natexlab{}.
\newblock \bibinfo{title}{{ShinkaEvolve}: Towards Open-Ended and
  Sample-Efficient Program Evolution}.
\newblock
\newblock
\showeprint[arxiv]{2509.19349}~[cs.CL]
\newblock
\shownote{arXiv:2509.19349}.


\bibitem[Lehman and Stanley(2011)]%
        {lehman2011novelty}
\bibfield{author}{\bibinfo{person}{Joel Lehman} {and}
  \bibinfo{person}{Kenneth~O. Stanley}.} \bibinfo{year}{2011}\natexlab{}.
\newblock \showarticletitle{Abandoning objectives: Evolution through the search
  for novelty alone}.
\newblock \bibinfo{journal}{\emph{Evolutionary Computation}}
  \bibinfo{volume}{19}, \bibinfo{number}{2} (\bibinfo{year}{2011}),
  \bibinfo{pages}{189--223}.
\newblock


\bibitem[Lewis et~al\mbox{.}(2017)]%
        {lewis2017deal}
\bibfield{author}{\bibinfo{person}{Mike Lewis}, \bibinfo{person}{Denis Yarats},
  \bibinfo{person}{Yann Dauphin}, \bibinfo{person}{Devi Parikh}, {and}
  \bibinfo{person}{Dhruv Batra}.} \bibinfo{year}{2017}\natexlab{}.
\newblock \showarticletitle{Deal or No Deal? End-to-End Learning of Negotiation
  Dialogues}. In \bibinfo{booktitle}{\emph{Proceedings of the 2017 Conference
  on Empirical Methods in Natural Language Processing}}.
  \bibinfo{publisher}{Association for Computational Linguistics},
  \bibinfo{address}{Copenhagen, Denmark}, \bibinfo{pages}{2443--2453}.
\newblock
\urldef\tempurl%
\url{https://doi.org/10.18653/v1/D17-1259}
\showDOI{\tempurl}


\bibitem[Li(2026)]%
        {li2026dictionaries}
\bibfield{author}{\bibinfo{person}{Pan Li}.} \bibinfo{year}{2026}\natexlab{}.
\newblock \bibinfo{title}{Dictionaries, Not Darwin: Set-Level Selection Beats
  {LLM} Evolution in Scientific Equation Discovery}.
\newblock
\newblock
\showeprint[arxiv]{2607.04108}~[cs.LG]
\newblock
\shownote{arXiv:2607.04108}.


\bibitem[Meyerson et~al\mbox{.}(2023)]%
        {meyerson2024lmx}
\bibfield{author}{\bibinfo{person}{Elliot Meyerson}, \bibinfo{person}{Mark~J.
  Nelson}, \bibinfo{person}{Herbie Bradley}, \bibinfo{person}{Adam Gaier},
  \bibinfo{person}{Arash Moradi}, \bibinfo{person}{Amy~K. Hoover}, {and}
  \bibinfo{person}{Joel Lehman}.} \bibinfo{year}{2023}\natexlab{}.
\newblock \bibinfo{title}{Language Model Crossover: Variation through Few-Shot
  Prompting}.
\newblock
\newblock
\showeprint[arxiv]{2302.12170}~[cs.NE]
\newblock
\shownote{arXiv:2302.12170}.


\bibitem[Moraglio et~al\mbox{.}(2012)]%
        {moraglio2012geometric}
\bibfield{author}{\bibinfo{person}{Alberto Moraglio},
  \bibinfo{person}{Krzysztof Krawiec}, {and} \bibinfo{person}{Colin~G.
  Johnson}.} \bibinfo{year}{2012}\natexlab{}.
\newblock \showarticletitle{Geometric Semantic Genetic Programming}. In
  \bibinfo{booktitle}{\emph{Parallel Problem Solving from Nature -- {PPSN}
  {XII}}} \emph{(\bibinfo{series}{Lecture Notes in Computer Science},
  Vol.~\bibinfo{volume}{7491})}. \bibinfo{publisher}{Springer},
  \bibinfo{address}{Berlin}, \bibinfo{pages}{21--31}.
\newblock
\urldef\tempurl%
\url{https://doi.org/10.1007/978-3-642-32937-1_3}
\showDOI{\tempurl}


\bibitem[Nolfi and Pagliuca(2024)]%
        {nolfi2025global}
\bibfield{author}{\bibinfo{person}{Stefano Nolfi} {and} \bibinfo{person}{Paolo
  Pagliuca}.} \bibinfo{year}{2024}\natexlab{}.
\newblock \showarticletitle{Global progress in competitive co-evolution: a
  systematic comparison of alternative methods}.
\newblock \bibinfo{journal}{\emph{Frontiers in Robotics and AI}}
  \bibinfo{volume}{11} (\bibinfo{year}{2024}).
\newblock
\urldef\tempurl%
\url{https://doi.org/10.3389/frobt.2024.1470886}
\showDOI{\tempurl}


\bibitem[Novikov et~al\mbox{.}(2025)]%
        {novikov2025alphaevolve}
\bibfield{author}{\bibinfo{person}{Alexander Novikov},
  \bibinfo{person}{Ng{\^a}n V{\~u}}, \bibinfo{person}{Marvin Eisenberger},
  \bibinfo{person}{Emilien Dupont}, \bibinfo{person}{Po-Sen Huang},
  \bibinfo{person}{Adam~Zsolt Wagner}, \bibinfo{person}{Sergey Shirobokov},
  \bibinfo{person}{Borislav Kozlovskii}, \bibinfo{person}{Francisco J.~R.
  Ruiz}, \bibinfo{person}{Abbas Mehrabian}, \bibinfo{person}{M.~Pawan Kumar},
  \bibinfo{person}{Abigail See}, \bibinfo{person}{Swarat Chaudhuri},
  \bibinfo{person}{George Holland}, \bibinfo{person}{Alex Davies},
  \bibinfo{person}{Sebastian Nowozin}, \bibinfo{person}{Pushmeet Kohli}, {and}
  \bibinfo{person}{Matej Balog}.} \bibinfo{year}{2025}\natexlab{}.
\newblock \bibinfo{title}{{AlphaEvolve}: A Coding Agent for Scientific and
  Algorithmic Discovery}.
\newblock
\newblock
\showeprint[arxiv]{2506.13131}~[cs.AI]
\newblock
\shownote{arXiv:2506.13131}.


\bibitem[Ouyang et~al\mbox{.}(2022)]%
        {ouyang2022training}
\bibfield{author}{\bibinfo{person}{Long Ouyang}, \bibinfo{person}{Jeffrey Wu},
  \bibinfo{person}{Xu Jiang}, \bibinfo{person}{Diogo Almeida},
  \bibinfo{person}{Carroll Wainwright}, \bibinfo{person}{Pamela Mishkin},
  \bibinfo{person}{Chong Zhang}, \bibinfo{person}{Sandhini Agarwal},
  \bibinfo{person}{Katarina Slama}, \bibinfo{person}{Alex Ray},
  \bibinfo{person}{John Schulman}, \bibinfo{person}{Jacob Hilton},
  \bibinfo{person}{Fraser Kelton}, \bibinfo{person}{Luke Miller},
  \bibinfo{person}{Maddie Simens}, \bibinfo{person}{Amanda Askell},
  \bibinfo{person}{Peter Welinder}, \bibinfo{person}{Paul Christiano},
  \bibinfo{person}{Jan Leike}, {and} \bibinfo{person}{Ryan Lowe}.}
  \bibinfo{year}{2022}\natexlab{}.
\newblock \showarticletitle{Training Language Models to Follow Instructions
  with Human Feedback}.
\newblock \bibinfo{journal}{\emph{Advances in Neural Information Processing
  Systems}}  \bibinfo{volume}{35} (\bibinfo{year}{2022}),
  \bibinfo{pages}{27730--27744}.
\newblock


\bibitem[Price(1970)]%
        {price1970selection}
\bibfield{author}{\bibinfo{person}{George~R. Price}.}
  \bibinfo{year}{1970}\natexlab{}.
\newblock \showarticletitle{Selection and Covariance}.
\newblock \bibinfo{journal}{\emph{Nature}} \bibinfo{volume}{227},
  \bibinfo{number}{5257} (\bibinfo{year}{1970}), \bibinfo{pages}{520--521}.
\newblock
\urldef\tempurl%
\url{https://doi.org/10.1038/227520a0}
\showDOI{\tempurl}


\bibitem[Pryzant et~al\mbox{.}(2023)]%
        {pryzant2023automaticpromptoptimizationgradient}
\bibfield{author}{\bibinfo{person}{Reid Pryzant}, \bibinfo{person}{Dan Iter},
  \bibinfo{person}{Jerry Li}, \bibinfo{person}{Yin~Tat Lee},
  \bibinfo{person}{Chenguang Zhu}, {and} \bibinfo{person}{Michael Zeng}.}
  \bibinfo{year}{2023}\natexlab{}.
\newblock \bibinfo{title}{Automatic Prompt Optimization with "Gradient Descent"
  and Beam Search}.
\newblock
\newblock
\showeprint[arxiv]{2305.03495}~[cs.CL]
\newblock
\shownote{arXiv:2305.03495}.


\bibitem[Rosin and Belew(1997)]%
        {rosin1997coevolution}
\bibfield{author}{\bibinfo{person}{Christopher~D. Rosin} {and}
  \bibinfo{person}{Richard~K. Belew}.} \bibinfo{year}{1997}\natexlab{}.
\newblock \showarticletitle{New methods for competitive coevolution}.
\newblock \bibinfo{journal}{\emph{Evolutionary Computation}}
  \bibinfo{volume}{5}, \bibinfo{number}{1} (\bibinfo{year}{1997}),
  \bibinfo{pages}{1--29}.
\newblock


\bibitem[Shinn et~al\mbox{.}(2023)]%
        {shinn2023reflexionlanguageagentsverbal}
\bibfield{author}{\bibinfo{person}{Noah Shinn}, \bibinfo{person}{Federico
  Cassano}, \bibinfo{person}{Edward Berman}, \bibinfo{person}{Ashwin Gopinath},
  \bibinfo{person}{Karthik Narasimhan}, {and} \bibinfo{person}{Shunyu Yao}.}
  \bibinfo{year}{2023}\natexlab{}.
\newblock \bibinfo{title}{{Reflexion}: Language Agents with Verbal
  Reinforcement Learning}.
\newblock
\newblock
\showeprint[arxiv]{2303.11366}~[cs.AI]
\newblock
\shownote{arXiv:2303.11366}.


\bibitem[Silver et~al\mbox{.}(2018)]%
        {silver2018general}
\bibfield{author}{\bibinfo{person}{David Silver}, \bibinfo{person}{Thomas
  Hubert}, \bibinfo{person}{Julian Schrittwieser}, \bibinfo{person}{Ioannis
  Antonoglou}, \bibinfo{person}{Matthew Lai}, \bibinfo{person}{Arthur Guez},
  \bibinfo{person}{Marc Lanctot}, \bibinfo{person}{Laurent Sifre},
  \bibinfo{person}{Dharshan Kumaran}, \bibinfo{person}{Thore Graepel},
  \bibinfo{person}{Timothy Lillicrap}, \bibinfo{person}{Karen Simonyan}, {and}
  \bibinfo{person}{Demis Hassabis}.} \bibinfo{year}{2018}\natexlab{}.
\newblock \showarticletitle{A general reinforcement learning algorithm that
  masters chess, shogi, and {Go} through self-play}.
\newblock \bibinfo{journal}{\emph{Science}} \bibinfo{volume}{362},
  \bibinfo{number}{6419} (\bibinfo{year}{2018}), \bibinfo{pages}{1140--1144}.
\newblock


\bibitem[Wang et~al\mbox{.}(2024)]%
        {wang2024awm}
\bibfield{author}{\bibinfo{person}{Zora~Zhiruo Wang}, \bibinfo{person}{Jiayuan
  Mao}, \bibinfo{person}{Daniel Fried}, {and} \bibinfo{person}{Graham Neubig}.}
  \bibinfo{year}{2024}\natexlab{}.
\newblock \bibinfo{title}{Agent Workflow Memory}.
\newblock
\newblock
\showeprint[arxiv]{2409.07429}~[cs.CL]
\newblock
\shownote{arXiv:2409.07429}.


\bibitem[Watson and Pollack(2001)]%
        {watson2001coevolutionary}
\bibfield{author}{\bibinfo{person}{Richard~A. Watson} {and}
  \bibinfo{person}{Jordan~B. Pollack}.} \bibinfo{year}{2001}\natexlab{}.
\newblock \showarticletitle{Coevolutionary dynamics in a minimal substrate}. In
  \bibinfo{booktitle}{\emph{Proc. Genetic and Evolutionary Computation Conf.
  ({GECCO})}}. \bibinfo{pages}{702--709}.
\newblock


\bibitem[Wu et~al\mbox{.}(2023)]%
        {wu2023autogenenablingnextgenllm}
\bibfield{author}{\bibinfo{person}{Qingyun Wu}, \bibinfo{person}{Gagan Bansal},
  \bibinfo{person}{Jieyu Zhang}, \bibinfo{person}{Yiran Wu},
  \bibinfo{person}{Beibin Li}, \bibinfo{person}{Erkang Zhu},
  \bibinfo{person}{Li Jiang}, \bibinfo{person}{Xiaoyun Zhang},
  \bibinfo{person}{Shaokun Zhang}, \bibinfo{person}{Jiale Liu},
  \bibinfo{person}{Ahmed~Hassan Awadallah}, \bibinfo{person}{Ryen~W White},
  \bibinfo{person}{Doug Burger}, {and} \bibinfo{person}{Chi Wang}.}
  \bibinfo{year}{2023}\natexlab{}.
\newblock \bibinfo{title}{{AutoGen}: Enabling Next-Gen {LLM} Applications via
  Multi-Agent Conversation}.
\newblock
\newblock
\showeprint[arxiv]{2308.08155}~[cs.AI]
\newblock
\shownote{arXiv:2308.08155}.


\bibitem[Yang et~al\mbox{.}(2024)]%
        {yang2024largelanguagemodelsoptimizers}
\bibfield{author}{\bibinfo{person}{Chengrun Yang}, \bibinfo{person}{Xuezhi
  Wang}, \bibinfo{person}{Yifeng Lu}, \bibinfo{person}{Hanxiao Liu},
  \bibinfo{person}{Quoc~V. Le}, \bibinfo{person}{Denny Zhou}, {and}
  \bibinfo{person}{Xinyun Chen}.} \bibinfo{year}{2024}\natexlab{}.
\newblock \bibinfo{title}{Large Language Models as Optimizers}.
\newblock
\newblock
\showeprint[arxiv]{2309.03409}~[cs.LG]
\newblock
\shownote{arXiv:2309.03409}.


\bibitem[Yao et~al\mbox{.}(2022)]%
        {yao2022react}
\bibfield{author}{\bibinfo{person}{Shunyu Yao}, \bibinfo{person}{Jeffrey Zhao},
  \bibinfo{person}{Dian Yu}, \bibinfo{person}{Nan Du}, \bibinfo{person}{Izhak
  Shafran}, \bibinfo{person}{Karthik Narasimhan}, {and} \bibinfo{person}{Yuan
  Cao}.} \bibinfo{year}{2022}\natexlab{}.
\newblock \showarticletitle{{ReAct}: Synergizing Reasoning and Acting in
  Language Models}.
\newblock \bibinfo{journal}{\emph{arXiv preprint arXiv:2210.03629}}
  (\bibinfo{year}{2022}).
\newblock


\bibitem[Yuan et~al\mbox{.}(2025)]%
        {yuan2025evoagent}
\bibfield{author}{\bibinfo{person}{Siyu Yuan}, \bibinfo{person}{Kaitao Song},
  \bibinfo{person}{Jiangjie Chen}, \bibinfo{person}{Xu Tan},
  \bibinfo{person}{Dongsheng Li}, {and} \bibinfo{person}{Deqing Yang}.}
  \bibinfo{year}{2025}\natexlab{}.
\newblock \showarticletitle{{EvoAgent}: Towards Automatic Multi-Agent
  Generation via Evolutionary Algorithms}. In
  \bibinfo{booktitle}{\emph{Proceedings of the 2025 Conference of the Nations
  of the Americas Chapter of the Association for Computational Linguistics:
  Human Language Technologies (Volume 1: Long Papers)}}.
  \bibinfo{publisher}{Association for Computational Linguistics},
  \bibinfo{address}{Albuquerque, New Mexico}, \bibinfo{pages}{6192--6217}.
\newblock
\urldef\tempurl%
\url{https://doi.org/10.18653/v1/2025.naacl-long.315}
\showDOI{\tempurl}


\bibitem[Zhang et~al\mbox{.}(2025)]%
        {zhang2025darwin}
\bibfield{author}{\bibinfo{person}{Jenny Zhang}, \bibinfo{person}{Shengran Hu},
  \bibinfo{person}{Cong Lu}, \bibinfo{person}{Robert Lange}, {and}
  \bibinfo{person}{Jeff Clune}.} \bibinfo{year}{2025}\natexlab{}.
\newblock \bibinfo{title}{Darwin G{\"o}del Machine: Open-Ended Evolution of
  Self-Improving Agents}.
\newblock
\newblock
\showeprint[arxiv]{2505.22954}~[cs.AI]
\newblock
\shownote{arXiv:2505.22954}.


\bibitem[Zhao et~al\mbox{.}(2024)]%
        {zhao2024expel}
\bibfield{author}{\bibinfo{person}{Andrew Zhao}, \bibinfo{person}{Daniel
  Huang}, \bibinfo{person}{Quentin Xu}, \bibinfo{person}{Matthieu Lin},
  \bibinfo{person}{Yong-Jin Liu}, {and} \bibinfo{person}{Gao Huang}.}
  \bibinfo{year}{2024}\natexlab{}.
\newblock \showarticletitle{{ExpeL}: {LLM} Agents Are Experiential Learners}.
  In \bibinfo{booktitle}{\emph{Proceedings of the {AAAI} Conference on
  Artificial Intelligence}}, Vol.~\bibinfo{volume}{38}.
  \bibinfo{pages}{19632--19642}.
\newblock
\urldef\tempurl%
\url{https://doi.org/10.1609/aaai.v38i17.29936}
\showDOI{\tempurl}


\end{thebibliography}

\end{document}